\documentclass[10pt,journal,compsoc]{IEEEtran}
\usepackage[caption=false,font=footnotesize]{subfig}
\usepackage{graphicx}
\usepackage{amssymb}
\usepackage{amsmath}
\usepackage{ctable}
\usepackage{arabtex}
\usepackage[T1]{fontenc}
\usepackage{utf8}
\setcode{utf8}

\begin{document}
\setlength{\parskip}{0.5em}

\title{A Sequence-to-Sequence Approach for Arabic Pronoun Resolution}

\author{Hanan S. Murayshid\textsuperscript{1,2}, 
        Hafida Benhidour\textsuperscript{1},
        Said Kerrache\textsuperscript{1}, 

\thanks{\\ 
    \textsuperscript{1} College of Computer and Information Sciences, King Saud University, Riyadh, Saudi Arabia \\ 
    \textsuperscript{2} King Abdulaziz City for Science and Technology - KACST, Riyadh, Saudi Arabia}}

\IEEEtitleabstractindextext{%
 \begin{abstract}
This paper proposes a sequence-to-sequence learning approach for Arabic pronoun resolution, which explores the effectiveness of using advanced natural language processing (NLP) techniques, specifically Bi-LSTM and the BERT pre-trained Language Model, in solving the pronoun resolution problem in Arabic. The proposed approach is evaluated on the AnATAr dataset, and its performance is compared to several baseline models, including traditional machine learning models and handcrafted feature-based models. Our results demonstrate that the proposed model outperforms the baseline models, which include KNN, logistic regression, and SVM, across all metrics. In addition, we explore the effectiveness of various modifications to the model, including concatenating the anaphor text beside the paragraph text as input, adding a mask to focus on candidate scores, and filtering candidates based on gender and number agreement with the anaphor. Our results show that these modifications significantly improve the model's performance, achieving up to 81\% on MRR and 71\% for F1 score while also demonstrating higher precision, recall, and accuracy. These findings suggest that the proposed model is an effective approach to Arabic pronoun resolution and highlights the potential benefits of leveraging advanced NLP neural models. 
\end{abstract}
\begin{IEEEkeywords}
Anaphora resolution, pronoun resolution, Arabic NLP,  sequence-to-sequence models, recurrent neural networks. 
\end{IEEEkeywords}
}
\IEEEdisplaynontitleabstractindextext
\maketitle
\IEEEpeerreviewmaketitle

\section{Introduction}
\IEEEPARstart{L}{anguage} cohesion is essential to human communication, and anaphora resolution is critical in achieving this cohesion. Anaphora resolution refers to identifying and resolving references to entities in a text. Within Natural Language Processing, anaphora resolution is a fundamental problem that has not received much attention, especially in Arabic.
One important type of anaphora resolution is pronoun resolution, which involves identifying and disambiguating the referents of pronouns in a text. Using pronouns is crucial to human language and vital in creating cohesive and comprehensible discourse. In written and spoken texts, pronouns are cohesive elements that link words, phrases, and sentences together to form a coherent whole. Pronoun resolution is an essential and challenging task, as pronouns are often ambiguous and can have multiple possible referents.
Correctly resolving pronouns impacts NLP applications, including information retrieval, text summarization, machine translation, and dialogue systems. Accurately resolving pronoun references is crucial for these applications to produce coherent and meaningful outputs.

Anaphora resolution techniques for the Arabic language are yet to receive their due attention from researchers. The current research landscape in the field highlights the need to employ state-of-the-art NLP techniques rather than relying on hand-crafted features. This paper aims to remedy this by investigating the ability of state-of-the-art NLP techniques to solve Arabic pronoun resolution. It introduces a novel sequence-to-sequence neural model for Arabic pronoun resolution that leverages pre-trained AraBERT \cite{antoun2020arabert} model as an encoder and a Bi-directional LSTM (Bi-LSTM)\cite{Bilstm-GRAVES2005602} as a decoder. The model produces a probability score for each token in the input sequence. The binary target sequence indicates whether each token is part of the correct antecedent. We compare the performance of the proposed model against existing linguistics hand-crafted features approaches on the AnATAr dataset \cite{hammamiArabicAnaphoraResolution2009}.

The rest of this paper is organized as follows. We first survey prior research on anaphora resolution techniques for the Arabic language and the application of neural network models in anaphora resolution in other languages. This is followed by a detailed description of the proposed model, including the input and target representations, the encoder and decoder architecture, and the training procedure. The paper also provides experimental results on the Arabic pronoun resolution dataset, AnATAr \cite{hammamiArabicAnaphoraResolution2009} corpus. Finally, the paper concludes by discussing the results and future work.

\section{Anaphora and Pronoun Resolution}
Anaphora resolution is a critical task in natural language processing that involves identifying the referent of an anaphoric expression, such as a pronoun, verb, or noun phrase in a given text. The phenomenon of anaphora is particularly challenging for Arabic, a Semitic language with complex morphological and syntactic structures. In Arabic, anaphoric expressions can refer to different kinds of antecedents, including nouns, pronouns, and verb phrases, and they can exhibit various grammatical and semantic relationships with their antecedents. Anaphora Resolution is typically observed in discourse, a formal mode of written or spoken communication consisting of multiple sentences or utterances collectively conveying a clear meaning and purpose \cite{Sukthanker2018, NemcikThesis2006}.

The coherence and cohesion of discourse are essential qualities that enable the reader to comprehend the semantic and cognitive meaning of the text and understand the connections between different parts of the text. According to Halliday and Hasan (1976) \cite{bernhardtReviewCOHESIONENGLISH1976}, cohesion is achieved through linguistic elements, such as anaphoric references, which establish links between different parts of the discourse. Anaphoric references can be exophoric and endophoric, depending on whether they refer to entities outside or within the discourse, respectively. Referring expressions identify individual objects in the discourse and can be noun phrases or surrogates for noun phrases. Anaphora, a particular type of endophora, involves relating an expression (anaphor) to another expression (antecedent) previously mentioned in the discourse. In natural language processing, anaphora resolution is a critical task that involves identifying the referent of an anaphoric expression, such as a pronoun, verb, or noun phrase, in a given text. Anaphora resolution has several key components, including the anaphor and antecedent, and the process of finding the antecedent that the anaphor refers to \cite{Sukthanker2018}. This process helps eliminate ambiguity in the meaning of the text and ensures the coherence and consistency of the text's context to understand its meaning better \cite{Bouzid2017, mathlouthibouzidAggregationWordEmbedding2019a}. Given these benefits, anaphora resolution is considered one of the most frequently used linguistic elements in natural language processing \cite{abolohomHybridApproachPronominal2015}.

\begin{figure}[!t]
  \centering
  \subfloat[Anaphor (highlighted) with candidate antecedents underlined.]{\fbox{\includegraphics[width=\linewidth]{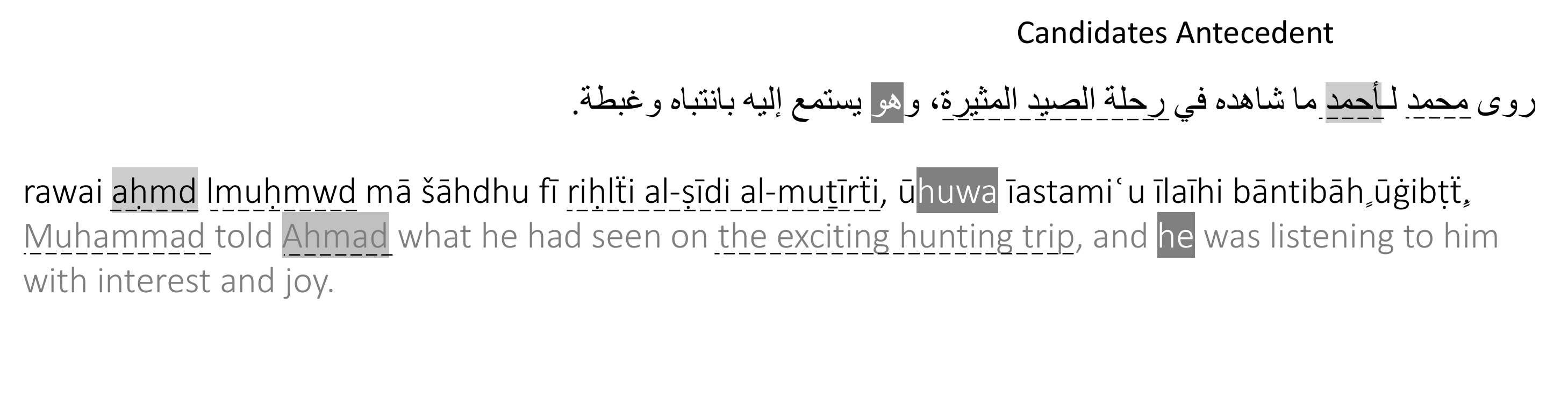}}%
  \label{fig:ex1a}}
  \hfil
   \subfloat[Anaphor with correct antecedent highlighted.]{\fbox{\includegraphics[width=\linewidth]{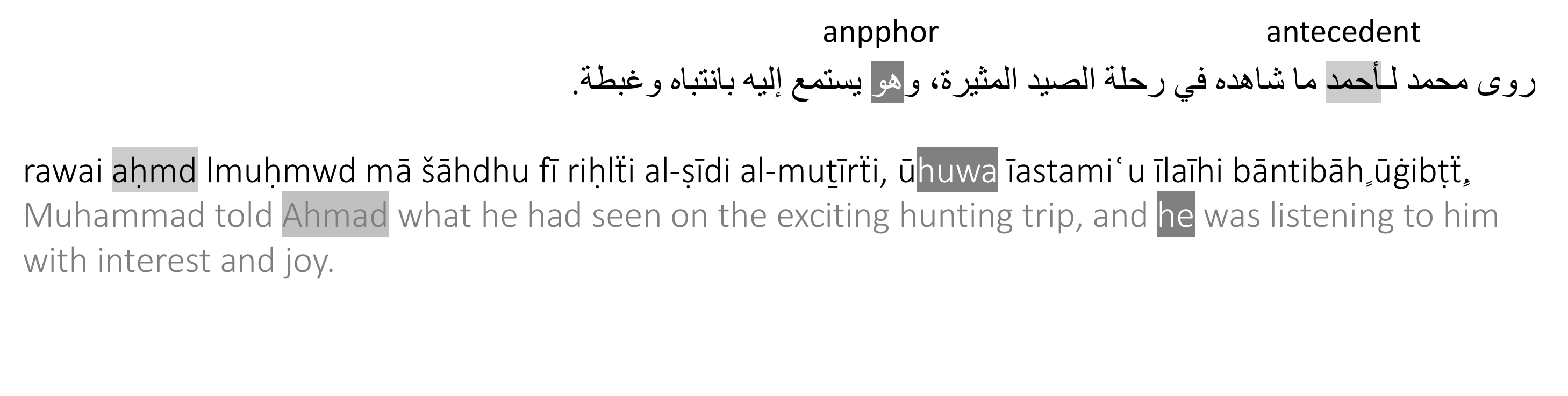}}%
  \label{fig:ex1b}}
  \caption{An example of Arabic pronoun resolution.}
  \label{fig:example}
\end{figure}

Pronoun resolution, a frequently used and well-known subcase of anaphora resolution \cite{Sukthanker2018, abolohomHybridApproachPronominal2015}, involves interpreting pronouns in text and identifying their corresponding antecedents, which can take various linguistic forms. Specifically, in the case of nominal anaphora resolution, the anaphor (pronoun) refers back to nouns or noun phrases. Pronoun resolution aims to resolve ambiguity and ensure an accurate text interpretation by finding the correct antecedent.
For instance, in Figure \ref{fig:example}, the pronoun/anaphor "\RL{هو}" (he) refers back to the noun antecedent "\RL{أحمد}" (Ahmed). The goal of pronoun resolution is to determine which one of the candidate antecedents, as shown in Figure \ref{fig:ex1a}, is the correct antecedent for the pronoun "\RL{هو}".

\subsection{Resolution Approaches}
Anaphora and pronoun resolution approaches can be classified in various ways. One common method categorizes them based on the resolution method, such as rule-based, corpus-based, and hybrid-based. Rule-based approaches rely on handcrafted rules to identify antecedents, whereas corpus-based ones use statistics and patterns from a corpus of data. Hybrid-based approaches combine both rule-based and corpus-based methods. They can also involve combining the outputs of different rule-based and corpus-based methods in a post-processing step or using a rule-based method to pre-process the data before applying a corpus-based method. The exact implementation of a hybrid-based approach can vary depending on the specific application and the available data.

Another strategy is categorizing anaphora resolution approaches based on the interaction between factors, such as serialization, weighting, or machine learning features approaches. The serialization approach prioritizes specific factors in a predetermined order, whereas the weighting approach assigns weights to the factors before selecting them. The machine learning features approach merges all factors into a single vector input without distinguishing or filtering them \cite{schiehlen2004optimizing}.

\subsection{Arabic Anaphora Resolution} 
Arabic anaphora resolution has a relatively short history compared to other languages. The first work in this direction was part of a multilingual study by \cite{mitkovMultilingualRobustAnaphora1998} in 1998. This study's resolution process was adapted from English, with only minor modifications. The method employs a part-of-speech tagger to pre-process texts and checks the input against agreement and several antecedent indicators. Each indicator assigns candidates scores, and the candidate with the highest aggregate score is returned as the antecedent. The evaluation of the approach for Arabic was performed on an Arabic manual script containing 190 anaphors, achieving a success rate of around 95\%.

Regarding data availability, Hammami et al. made a crucial contribution to Arabic anaphora resolution in 2009 by constructing the first corpus, named AnATAr. This corpus represents a crucial step in developing annotated corpora for Arabic anaphora resolution \cite{hammamiArabicAnaphoraResolution2009}. In addition to AnATAr, another corpus named QurAna \cite{sharafQurAnaCorpusQuran2012} was established in 2012 using the Holy Quran as its base. This corpus was annotated with pronominal anaphora information to advance the Arabic anaphora resolution study.

In 2015, efforts were made to enhance the QurAna corpus by combining it with the QAC corpus \cite{SeddikArabicAnaphoraResolution2015} to obtain morphological features of Quranic segments. This work aimed to improve the level of annotation in the Holy Quran text for anaphora resolution and to provide morphological features for referential links in pronominal anaphora resolution.

In 2007, Elghamry et al. \cite{elghamryArabicAnaphoraResolution2007} presented the first work to resolve Arabic anaphora using machine learning, specifically supervised learning. The authors utilized statistical and dynamic machine learning approaches, employing conditional probability as the association measure to classify potential anaphor-candidate pairs. The performance of the proposed solution was evaluated using a web-based corpus built with a cues-based algorithm that relied on gender, number, and rationality cues extracted from the Arabic language. The proposed solution demonstrated a performance of 87.6\% when evaluated against a gold standard benchmark of 5000 pronouns.

From 2014, with the rise in popularity of Deep Learning, research in Arabic anaphora resolution continued to employ both statistical machine learning algorithms \cite{abolohomMachineLearningApproach2014} and a hybrid approach incorporating linguistic rules \cite{abolohomHybridApproachPronominal2015}. The classification problem was approached using algorithms such as Naive Bayes, K-nearest neighbors (KNN), and linear logistic regression. The proposed solutions were evaluated using the QurAna corpus, explicitly focusing on pronominal anaphors. They were framed as a classification task for each potential anaphor-candidate pair represented by preceding NP phrases.

In Mathlouthi and Bouzid's study \cite{mathlouthibouzidNovelApproachBased2016}, published in 2016, a novel approach to Arabic anaphora resolution was presented that deviates from the existing solutions. This approach is based on Reinforcement Learning, which utilizes the principles of rewards and punishments to learn from past experiences and make decisions using Markov Decision Process to resolve pronominal anaphora. The authors aim to address the limitations of traditional rule-based and machine-learning methods through Reinforcement Learning, which is well-suited to handle complex and dynamic situations. The AnATAr corpus evaluated the proposed solution, which considered only personal, demonstrative, and relative pronouns. The results showed a success rate of up to 83.33\%.

In addition, Bouzid and colleagues (2017, 2019, 2020, 2021) \cite{Bouzid2017,mathlouthibouzidAggregationWordEmbedding2019a, Bouzid2020}, and Trabelsi et al. (2018) \cite{trabelsiArabicAnaphoraResolution2018} have employed Reinforcement Learning to tackle the problem of anaphora resolution in Arabic texts. Their work utilized Markov Decision Process with different combinations of syntactic knowledge features obtained through Q-learning methods and semantic knowledge features derived using embedding models such as Word2Vec and BERT. The authors evaluated their approach using the AnATAr dataset and reported success rates of up to 72.03\%, 66.49\%, 79.37\%, 80\%, and 83.33\% in their respective studies.

In 2017, Abolohom and Omar \cite{abolohomComputationalModelResolving2017} proposed a rule-based solution based on linguistic rules, which goes through preprocessing the input using a POS tagger, chunker, and Grammatical Relation Identifier. They then identify anaphors by eliminating non-anaphoric pronouns and identifying the candidate antecedents for each anaphor. After checking number and gender agreements, they score the candidates based on linguistic rules to find the most probable antecedent. Their model achieved 84\% accuracy on the Quran Corpus.

In 2021, Abolohom et al. \cite{abolohomComparativeStudyLinguistic2021} collected a set of linguistic and computational features to evaluate their effectiveness in different combinations using various machine learning classifiers, such as K-nearest classifiers (KNN), decision trees, meta-classification, and maximum entropy. The authors conducted random experiments with varying combinations of features to determine which combination produced the best results. Their experiments showed that the meta-classifier gives better results with different feature combinations. However, the study did not summarize the results or identify the most important features based on their experiments.

As of the time of writing, the most recent contribution to the field of Arabic anaphora resolution was in 2021 \cite{cheraguiA3CArabicAnaphora2021}, which resulted in the creation of the Arabic Anaphora Annotated Corpus (A3C). This corpus was produced as part of a master's thesis and featured an anaphoric annotating tool that utilizes linguistic and statistical rules to identify anaphors and their referents automatically. The A3C corpus offers an annotated collection of pronominal and verbal anaphora and their corresponding referents, spanning various text genres. In addition to these links, the dataset provides valuable morphological and statistical information about both the referent and the anaphora, including gender, number, and frequency.

\subsection{Neural Network-based Pronoun Resolution in other Languages}
Using neural networks to capture informative information for reference resolution began early for Japanese and Chinese languages \cite{chen-ng-2016-chinese, iida-etal-2016-intra-japanes}, specifically for zero pronoun resolution, in which the pronoun is explicitly omitted. In \cite{yin-etal-2018-zero}, the authors proposed a solution that uses Recurrent Neural Network (RNN) and attention mechanisms to model zero-pronoun resolution. They treated it as a pairing of the context's most informative part and the candidate's antecedent. They then provided a resolution score with handcrafted features, selecting the pair with the highest score as the antecedent. Their training set comprised 1391 documents with 12K anaphoric zero pronouns, and they used an additional 2K documents for testing. Their model achieved an overall F-score of 57\%.

Subsequently, neural networks were utilized in anaphora and pronoun resolution for Asian languages, such as Telugu \cite{asian-bs-1911-09994}, achieving an F-score of 86\%. It is worth noting that their solution included handcrafted features such as gender-number agreement, the pronoun's person (i.e., first, second, or third), and part-of-Plural, in addition to embedded context. The problem was also modeled as anaphor-candidate pairs. For the Uyghur language \cite{Uyghur-9076012}, a solution for pronoun resolution was proposed that utilizes an Independently Recurrent Neural Network \cite{indrnn-abs-1803-04831} and a multi-level attention mechanism, taking into account word vectors, position (distance), and part-of-speech tags. Their dataset comprised 341 documents with 38K examples for training and 86 documents with 6K examples for testing. This approach achieved an F-score of 83.8\%. Lastly, in the context of neural network-based anaphora resolution, the authors in \cite{liu-etal-2019-referential} proposed a solution mimicking human interpretation by sequentially reading the text from left to right and storing entities in memory. This network, termed the Recurrent Entity Network, utilizes Gated Recurrent Unit \cite{choetal2014learning} for the recurrent unit and a memory unit for updating and storing entities. They tested their solution on a gender-balanced dataset (GAP) \cite{webster2018gap} consisting of 8K examples. Alongside BERT \cite{devlin-etal-2019-bert}, they achieved an F-score of 78\%.

\section{Proposed Solution} 
All previous solutions for Arabic anaphora resolution have relied on handcrafted features, rules, and criteria based on linguistic perspectives. Despite the existence of Deep Learning techniques for various NLP tasks since 2014, such as Convolutional Neural Networks (CNNs), Recurrent Neural Networks, and Transformer models, which replaced handcrafted features by learning representations, previous Arabic anaphora resolution solutions have not incorporated such techniques. The sequence-to-sequence learning paradigm, which processes the entire sequence simultaneously or sequentially, has also shown great promise for various NLP tasks, including named entity recognition and machine translation. Therefore, we propose to explore the use of Deep Learning techniques on NLP and the sequence-to-sequence learning paradigm for Arabic anaphora resolution.

\subsection{Problem Formulation} 
How a problem is formulated can significantly impact the techniques available to solve it and the resulting performance. In order to specify a problem, three main elements need to be considered: the input format, the task, and the output format. Anaphora resolution typically involves several steps, including identifying a set of anaphors, identifying a set of candidate antecedents for each anaphor, and determining the correct antecedent.

Prior to the advent of machine learning-based approaches, the task of Arabic anaphora resolution was commonly tackled through the use of preference filtering, wherein a set of linguistic rules and agreements were applied to filter a list of candidates, as employed in \cite{mitkovMultilingualRobustAnaphora1998} and \cite{abolohomComputationalModelResolving2017}. With the emergence of machine learning techniques, the problem has been framed as a binary classification task, where mention-pair anaphor with a candidate antecedent serve as input, and the output indicates whether the pair is correct or not as employed in \cite{abolohomMachineLearningApproach2014} and \cite{abolohomHybridApproachPronominal2015}. Each pair representing an anaphor with a correct antecedent then gives rise to $k$ examples: one positive and $k-1$ negative examples, where $k$ is the number of candidates antecedent for the anaphor. This formulation leads to an unbalanced dataset with a small number of positives and many negatives, which can be addressed by limiting the negative candidates. However, in low-resource languages such as Arabic, this can result in a small dataset that requires augmentation. Additionally, it can be challenging to generate reliable anaphoric text for the learning phase on linguistically complex tasks such as anaphora resolution.

Other research efforts in Arabic anaphora resolution, such as those proposed in \cite{mathlouthibouzidNovelApproachBased2016, Bouzid2017, mathlouthibouzidAggregationWordEmbedding2019a, Bouzid2020}, as well as \cite{trabelsiArabicAnaphoraResolution2018}, incorporate reinforcement learning techniques. These approaches frame the task as selecting the most appropriate antecedent from a list of candidates for a given anaphor based on linguistic criteria. The reinforcement learning algorithm is employed to identify the optimal combination of criteria that can most effectively distinguish the correct antecedent from the other candidates.

In this work, we formulate the pronoun resolution problem in a way that facilitates using sequence-to-sequence models. Formally, let $P$ be a paragraph consisting of the sequence of words $w_1, w_2, \ldots, w_n$, $n$ being the number of words in the paragraph. Depending on the tokenization scheme, each word $w_i$ may get segmented into one or multiple tokens $t^i_1, t^i_{2}, \ldots, t^i_{l_i}$, where $l_i$ is the length of the word $w_i$ in terms of tokens. Therefore, after tokenization, paragraph $P$ will be represented as a sequence of tokens $t_1, t_2, \ldots t_m$, where $m$ is the total number of tokens in the paragraph. 
The task of the pronoun resolution algorithm is to assign a score $s_{i} \in [0,1]$ to each token $t_{i}$ in $P$, indicating the likelihood that it is the correct antecedent for the anaphoric expression, which can be a word (detached pronoun) or part of a word (attached pronoun) that refers to some entity mentioned earlier in $P$ (the antecedent).

\subsection{Sequence-to-sequence Models}
Sequence-to-sequence \cite{seq2seq-10.5555/2969033.2969173} models are a class of neural network models that can map an input sequence to an output sequence, even if the two sequences have different lengths or belong to different domains, such as translating between two languages. While initially developed for machine translation and speech recognition, these models are now widely used in various natural language processing tasks, such as text summarization and dialogue generation.

One of the earliest and most well-known Sequence-to-sequence models is the encoder-decoder architecture, proposed in 2014 by Cho et al. \cite{choetal2014learning}. In this architecture, the encoder converts the input sequence into a fixed-length vector representation while the decoder generates the output sequence based on the encoded vector. The encoder and decoder can be implemented using Recurrent Neural Networks, but Long short-term memory (LSTMs) or bidirectional LSTMs are commonly used. The encoder-decoder architecture has served as the foundation for many subsequent Sequence-to-sequence models. The success of the encoder-decoder model can be attributed to its ability to encode the input sequence into a dense vector representation, effectively capturing long-term dependencies between elements in the sequence. In addition, the model's decoder can generate output sequences of variable length, making it highly flexible for various applications.

\subsection{Model Architecture}
To each token $t_{i}$ in paragraph $P$, we assign an embedding vector $v_{i} \in \mathbb{R}^d$, where $d$ is the dimensionality of the embedding space, produced by a pre-trained language model, such as AraBERT \cite{antoun2020arabert}. The input vector $x_{i}$ corresponding to the token $t_{i}$ is the concatenation of its embedding vector $v_{i}$ and two binary features indicating whether the token is an anaphoric expression and a candidate expression:
\begin{equation}
x_i = [v_i; z_i; c_i],
\label{eq:x_1}
\end{equation}
where
\begin{equation}
z_i = Z(t_i) = \begin{cases}
1 & \text{if } t_{i} \in \text{ an anaphoric expression}, \\
0 & \text{otherwise},
\end{cases}
\end{equation}
and
\begin{equation}
c_i = C(t_i) = \begin{cases}
1 & \text{if } t_{i} \in  \text{ a candidate expression}, \\
0 & \text{otherwise}.
\end{cases}
\end{equation}
The paragraph is therefore represented by the input sequence $X = (x_1, x_2, \ldots, x_m)$. The target sequence, which has the same length as $X$, is denoted by $Y = (y_1, y_2, \ldots, y_m)$, where $y_i \in \{0, 1\}$ indicates whether the $i$-th token in $P$ is part of the correct antecedent. Namely, $y_i=1$ if token $t_i$ belongs to the antecedent and $y_i=0$ otherwise.

\begin{figure*}[!t]
 \centering
 \includegraphics[width=0.9\textwidth]{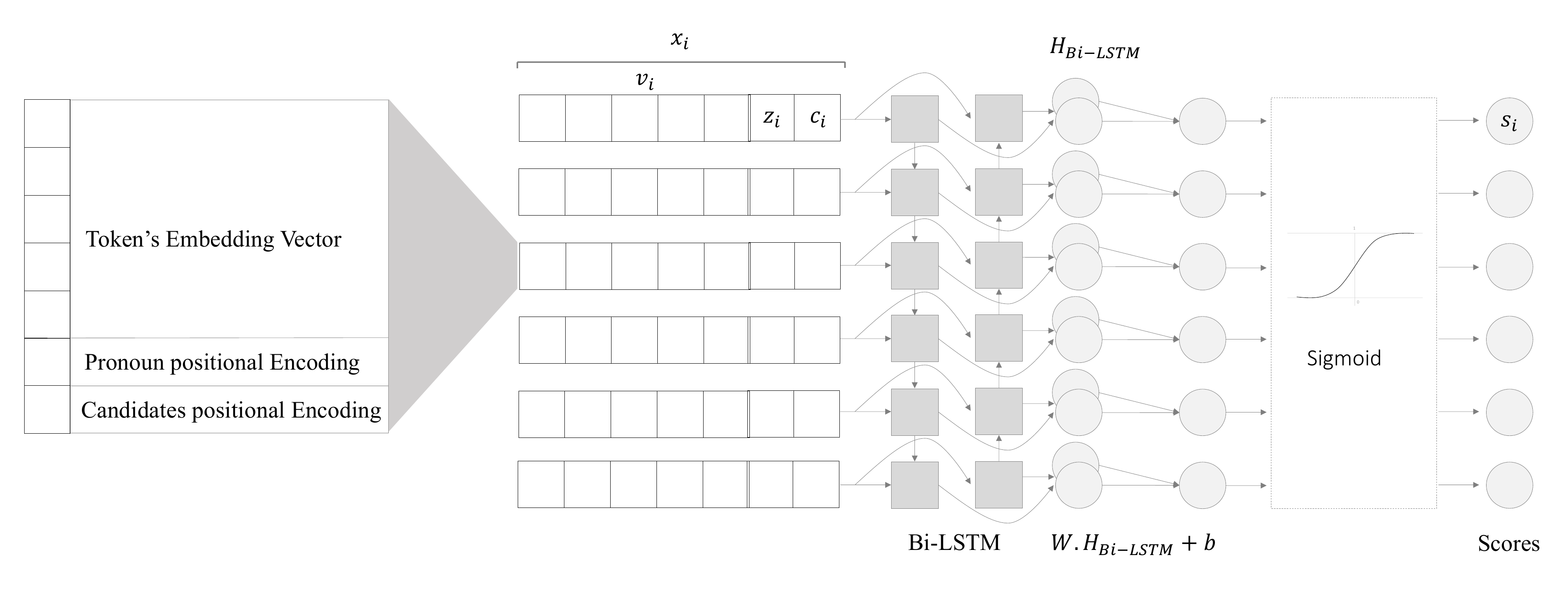} 
 \caption{The architecture of the proposed model.}
 \label{fig:model} 
\end{figure*}

As shown in Figure \ref{fig:model}, the proposed model has an encoder-decoder architecture consisting of a pre-trained Arabic BERT encoder and a Bi-LSTM network as decoder.
The encoder tokenizes the input paragraph $P = (w_1, w_2, \ldots, w_n)$ and produces a sequence of token embeddings $v_1, v_2, \ldots, v_m$, which are then concatenated with binary features to produce the input sequence $X = (x_1, x_2, \ldots, x_m)$ as described in Eq. \eqref{eq:x_1}.
The decoder, on the other hand, consists of a Bi-LSTM layer that takes the sequence $X$ as input and produces a sequence of scores $S = (s_1, s_2, \ldots, s_m)$, which represent the probability that each token is a part of the correct antecedent. Bi-LSTM is a recurrent neural network capable of learning long-term dependencies in a sequence. It consists of two LSTM layers, one processing the sequence forwards and the other processing it backward, enabling the model to capture information from past and future contexts. The output of the Bi-LSTM layer is obtained by concatenating the forward and backward hidden states:
\begin{equation}
H_{Bi-LSTM} = [H_{LSTM}^f;H_{LSTM}^B]
\end{equation}
defined as:
\begin{align}
H_{LSTM}^f &= (h_1, h_2, \ldots, h_{n-1}, h_n),\\
H_{LSTM}^b &= (h_n, h_{n-1}, \ldots, h_2, h_1),
\end{align}
where $h_i$ is the hidden state of the LSTMs upon processing the $i$-th element of the sequence, $H_{LSTM}^f$ and $H_{LSTM}^b$ are the sequences of hidden states of the forward and backward LSTM respectively.

The decoder's hidden states $H_{Bi-LSTM}$ are then passed through a linear layer followed by a sigmoid function to obtain the sequence of output scores $S$:
\begin{equation}
S = \sigma \left(W \cdot H_{Bi-LSTM} + b \right).
\end{equation}
Using the sigmoid activation function allows the model to handle the case of antecedents having multiple tokens resulting in output sequences having multiple entries set to 1, unlike the softmax function, which forces the sum of all scores to be 1 instead, making it more suitable for cases where only one of the outputs is 1 with the rest being 0.

To train this model, we minimize the binary cross-entropy loss between the predicted scores and sequence labels. The target sequence consists of binary values of 1 for the tokens belonging to the correct antecedent and 0 for the other tokens. The aim is to maximize the likelihood between the predicted scores (probability of being the correct antecedent) and the binary labels. The binary cross-entropy loss, also known as binary log loss or logistic loss, is commonly used in binary classification problems, where the targets are either of two classes (often represented as 0 and 1). It quantifies the difference between two probability distributions: the true distribution (the actual labels in your data) and the estimated distribution (the predictions from your model) \cite{murphy2012machine}. It is formally defined as follows:
\begin{equation}
\text{BCELoss} = -\frac{1}{N} \sum_{i=1}^N y^i \log \hat{y}^i + (1 - y^i) \log (1 - \hat{y}^i),
\end{equation}
where $y$ is the target value (binary values of 0 or 1), $\hat{y}$ is the predicted value (value between 0 and 1), and $N$ is the number of training examples. The loss function calculates the difference between the predicted probabilities and the true values and penalizes the model for incorrect predictions.

In our case, we calculate the loss over the entire sequence rather than one element for every example:
\begin{equation}
\text{Loss} = -\frac{1}{N}\sum_{i=1}^{N}\sum_{j=1}^{m}\left[y_j^i\log(s_j^i) + (1-y_j^i)\log(1-s_j^i)\right],
\end{equation}
where $N$ is the number of training examples, $m$ is the length of the target sequence, $y_j^i$ is the true value of the $j$-th token of the target sequence for the $i$-th training example, and $s_j^i$ is the predicted score of the $j$-th token of the target sequence for the $i$-th training example.

\subsection{Variants}
To explore more in-depth the capabilities of the proposed approach, we introduce various variants of the base model, each building on the strengths of the previous one:
\begin{itemize}
\item Appending anaphor text: The first modification is to append the anaphor text to the end of the paragraph text, creating a single input sequence for the model. This adjustment is based on the observation that the anaphor is often related to the preceding context in the paragraph and that including this context may enhance the model's ability to identify the correct referent. To preserve the context of the anaphor text during input encoding, we first embed the paragraph, then extract the anaphor portion of the embedding and attach it to the end of the sequence.

\item Masking non-candidate tokens: The second modification we make is to add a mask after the sigmoid layer to focus solely on candidates, which attributes more significant weight to candidate tokens and improves the model's ability to identify the most likely referent for each anaphor. The mask is implemented by multiplying the output of the sigmoid layer by a binary mask vector. The mask vector had a value of 1 for each candidate that met the nominal criteria and a value of 0 for all other tokens:
\begin{equation}
M(S, mask) = S \odot mask + \epsilon,
\label{eq:candidatesmask}
\end{equation}
where $S$ is the vector of scores at the output of the sigmoid layer, $\odot$ is element-wise product,  and $\epsilon$ is a small positive number added to avoid numerical problems when computed the derivative of the loss (in practice, we used $\epsilon= 1e-16)$.

\item Filtering candidates based on agreement: The last modification we introduce is incorporating additional criteria into the candidate selection process. In addition to the nominal criteria used in the base model, we filter the candidates based on gender and number agreement with the anaphor. This modification aims to improve the model's performance by reducing confusion between referents.
\end{itemize}

\section{Experimental Evaluation}
This section includes a detailed description of the experimental design, including the dataset used and how it was processed, and the competing models. The section further describes how the training process was used to optimize the model and the evaluation metrics used to compare its performance to that of the competing models.

\subsection{Dataset}
The dataset used in this study is the AnATAr corpus \cite{hammamiArabicAnaphoraResolution2009}, a publicly available dataset for Arabic Anaphora Resolution, which includes various types of anaphora, including pronouns. The corpus was first published in 2009 and consists of text documents from Tunisian educational books annotated with "EXP" and "PTR" tags for anaphors and their antecedents.

\subsubsection{Data Format Conversion}
\begin{figure}[!t]
 \centering
 \includegraphics[width=\columnwidth]{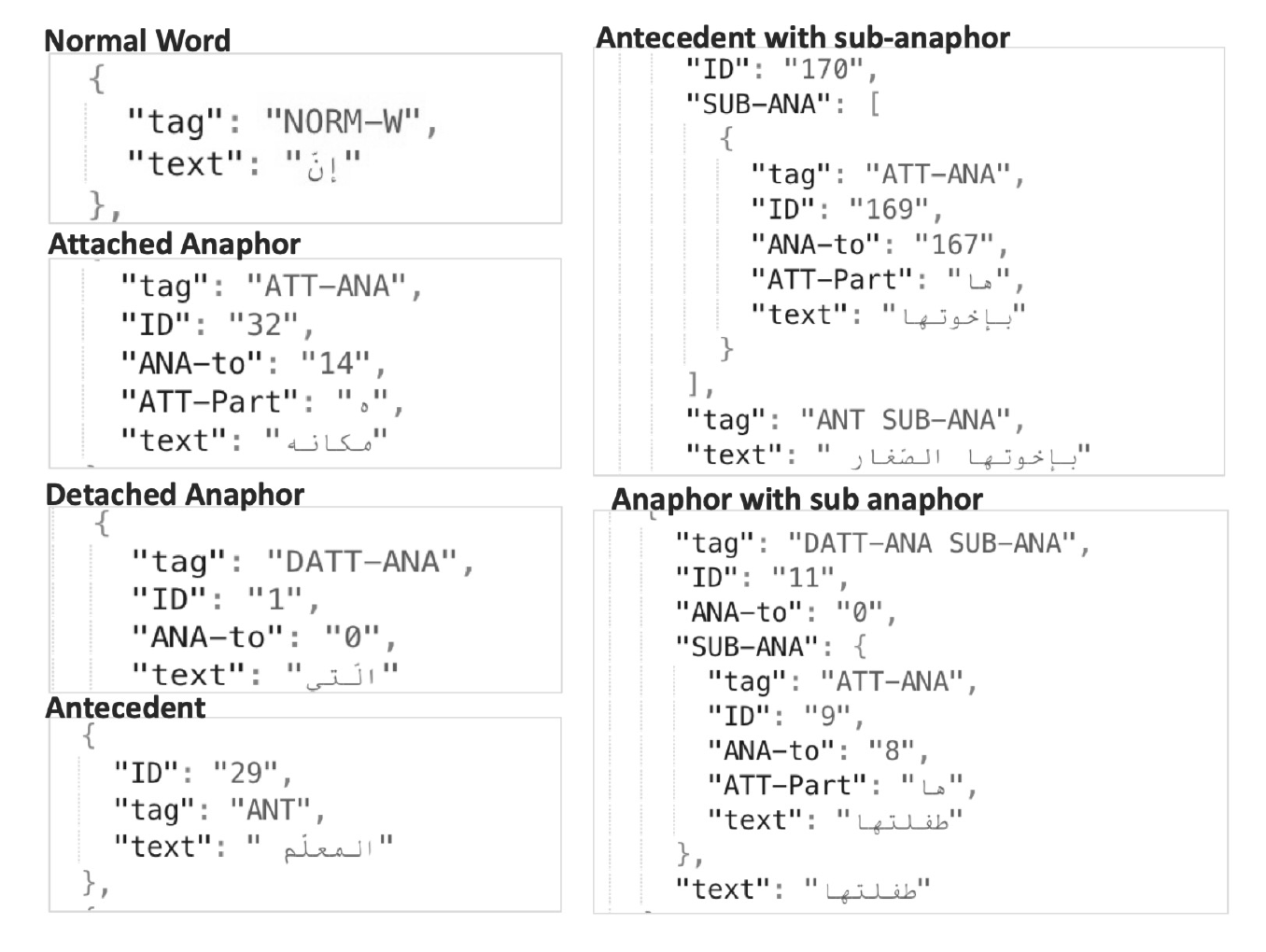} 
 \caption{Examples from the reformated data stored in JSON format.}
 \label{fig:jsondata} 
\end{figure}
To streamline the data processing workflow, we first addressed the complexity of the annotated documents, which are formatted in an XML  hierarchical structure. We transformed the XML documents into a sequential JSON format, representing each sentence as a list of words. Each word contains pertinent information, such as its part-of-speech tag, whether an anaphor, antecedent, or ordinary word. We assign an ID extracted from the original corpus to each antecedent, and each anaphor contains the ID of the antecedent to which it refers. Figure \ref{fig:jsondata} shows some examples of the reformated data.

\subsubsection{Data Cleaning}
Preparing the data for natural language processing requires careful cleaning to remove noise and irrelevant information. In this work, we performed several data-cleaning procedures to ensure the dataset's quality. The first step was to eliminate relations without antecedents caused by annotation errors. Additionally, relations with chain nature were simplified by selecting the nearest noun antecedent, as this study focused on anaphora resolution (pairs), not coreference resolution (clusters). Finally, the dataset was restricted to pronominal anaphoric expressions, which accounted for approximately 88\% of the data. This decision was based on the prevalence of pronominal anaphors in Arabic and the focus of many studies in the field \cite{abolohomHybridApproachPronominal2015, beseisoCoreferenceResolutionApproach2016, abolohomComputationalModelResolving2017, Bouzid2017, mathlouthibouzidAggregationWordEmbedding2019a}. Applying these data-cleaning procedures ensured a high-quality dataset suitable for training and evaluating the proposed model.

\subsubsection{Identifying Candidate Antecedents}
In this study, we focus on the pronoun and nominal anaphora resolution; thus, the candidate list is restricted to nouns and noun phrases. The AnATAr corpus used in this study is already segmented into word items, which may be standalone words or phrases, and we have followed this original segmentation. To determine whether a segment is a candidate antecedent, we have used CAMeL \cite{obeidetal2020camel} along with CoreNLP Stanford \cite{corenlpStanfordTagger2003} taggers as a double-check to identify the candidate word or phrase. The search window for candidates is open from the beginning of the document until the sentence containing the anaphoric expression (pronoun) is encountered. This approach ensures that all potential antecedents are identified and considered during the resolution process.

\subsection{Competing Models}
Since no previous sequence-to-sequence solutions for Arabic pronoun resolution exist in the literature, we reformulated exisitng traditional machine learning methods based on handcrafted features into a sequence-to-sequence format. We employed several algorithms, including K-Nearest Neighbors (KNN) \cite{1053964-knn}, Support Vector Machine (SVM) \cite{10.1023/A:1022627411411-svm}, and Logistic Regression \cite{cox1958regression} to perform inference as sequence-to-sequence models.

\begin{figure}[!t]
 \centering
 \includegraphics[width=\columnwidth]{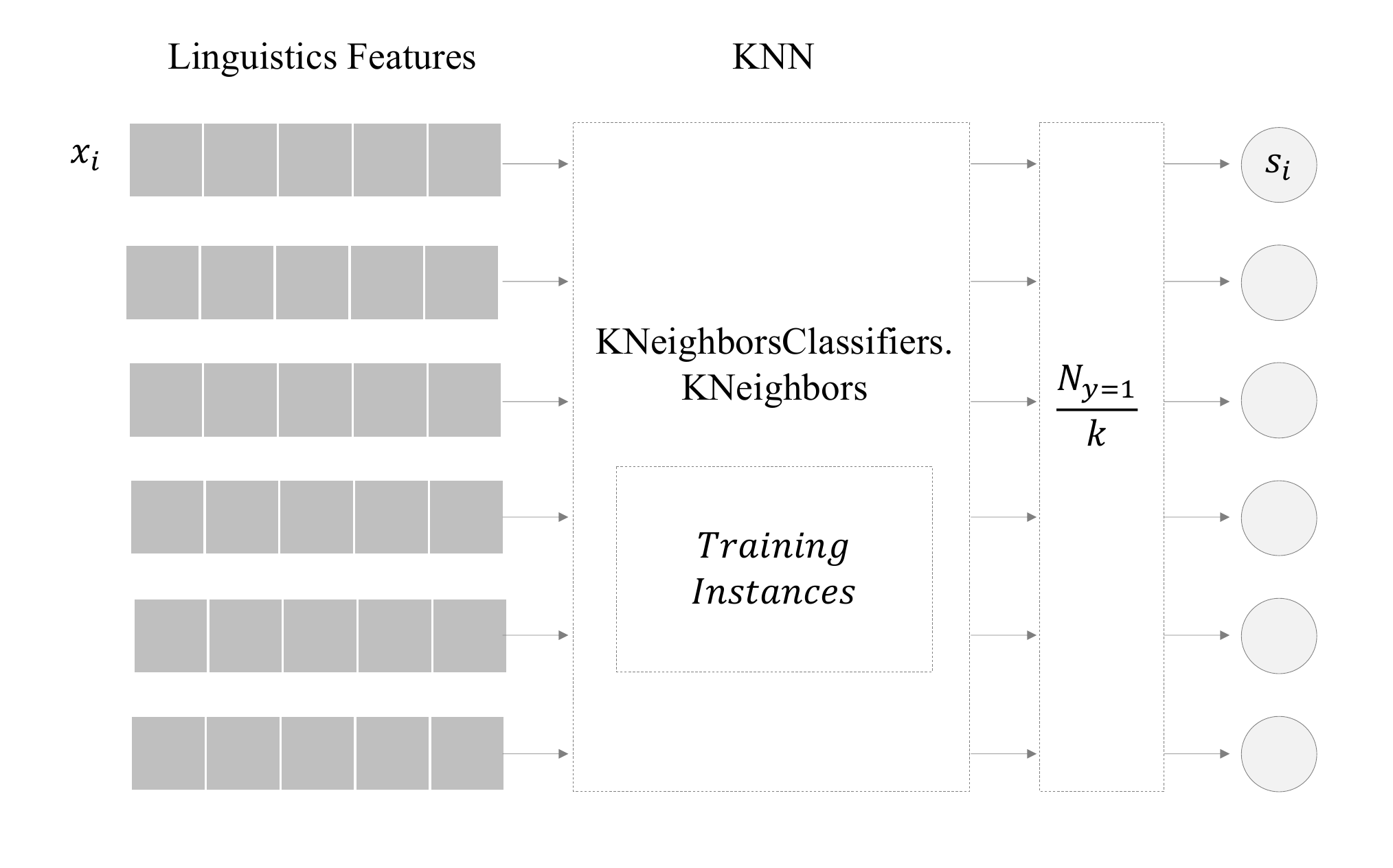} 
 \caption{The architecture of the KNN model used in performance evaluation.}
 \label{fig:baseline} 
\end{figure}

We treated the input sequence text as a sequence of tokens and encoded each token as a vector of basic linguistic features. These features included number agreement and gender agreement between the word containing the token and the anaphor (pronoun), whether it was part of a definite noun phrase, the sentence-based distance between the token and the anaphor (pronoun), and the person of the pronoun (i.e., first, second, or third).
For KNN, we first identified the $k$-nearest neighbors to the token's feature vector and then calculated the probability of the neighbors being predicted as antecedents for the pronoun (i.e., with a label of 1). The word with the tokens' highest probability (score) value was selected as the antecedent, as illustrated in Figure \ref{fig:baseline}.

The SVM approach works by finding an optimal hyperplane that maximizes the margin between classes. As with KNN, we calculated the probability of a token's features being classified as class 1, using a function implemented by the scikit-learn library \cite{scikit-learn}. The word with the highest probability value among the tokens was selected as the antecedent for the given example. The same procedure was applied for Logistic Regression, a type of generalized linear model (GLM) that models the relationship between predictor variables and a binary outcome using the logistic function \cite{0c96dc6df2c940d5a61876729acdbcb7}.

\subsection{Performance Measures}
The performance of the proposed pronoun resolution model is evaluated using several metrics, including Mean Reciprocal Rank (MRR), accuracy, precision, recall, and F1 score. MRR is commonly used in information retrieval and question answering tasks to measure the rank of the correct answer among a set of candidate answers \cite{QA-IR-MM}. In the context of our approach, the model outputs a score for each token in the input sequence, and the correct antecedent is expected to have the highest score. Therefore, we use MRR to measure how well the model ranks the correct antecedent relative to others by getting the max tokens' scores that build the words or phrases based on the given tokenized text. A perfect model would have an MRR of 1, indicating that the correct antecedent is ranked first in all cases. MRR@1 is defined as:
\begin{equation}
MRR = \frac{1}{|T|} \sum_{i=1}^{|T|} \frac{1}{rank(ant_i)},
\end{equation}
where $|T|$ is the number of examples in the test set $T$, and $rank(ant_i)$ is the rank of the correct antecedent for the $i$-th example. In addition to MRR, we report accuracy, precision, recall, and F1 measure.

Whereas the MRR measures the rank of the correct antecedent, the other metrics measure the prediction success for each token. Since the token classes are unbalanced, accuracy can be misleading and may not reflect the model's actual performance; The model may perform well on the majority class but poorly on the minority class, which in our case, is the more important class to identify correctly. In contrast, the F1 score is often considered the best choice for unbalanced classes, as it balances both precision and recall and considers false positives and false negatives. Additionally, precision measures how often the model correctly identifies the positive class. These metrics comprehensively evaluate the model's Arabic pronoun resolution task performance.

\subsection{Experimental Setup}
We split the AnATAr dataset into training and testing sets with a 70:30 ratio. More specifically, we divided the dataset at the document level using 70\% of the 59 documents for training and the remaining 30\% for testing. This approach ensures non-redundancy in the text between the two sets and that multiple examples represent every document. As discussed in the previous section, the dataset underwent preprocessing procedures before being used for experimentation.

We fitted our model and the competing models using the same training set. We used the Adam optimizer to train our model with a learning rate of 0.005 and a batch size of 16. We trained the model for a maximum of 20 epochs and applied the early stopping technique with patience of 5 epochs.
The competing models are all implemented using the scikit-learn library. To determine the optimal number of neighbors $k$ for the KNN algorithm, implemented by the class \verb|KNeighborsClassifier|, we performed a hyperparameter search using \verb|GridSearchCV| from the scikit-learn library. We varied $k$ from 10 to 30 and selected the value that achieved the highest performance. The scikit-learn implementation of KNN was used and fitted to the training data. To obtain the $k$ neighbors, we utilized the \verb|kneighbors| function and calculated the probability of a label equal to 1 among the neighbors. Regarding SVM and Logistic Regression, we used the default settings provided by the scikit-learn library.

\subsection{Results and Discussion}

\begin{table*}[!t]
    \centering
    \caption{Performance results obtained on the AnATAr dataset. The best result for each performance measure is highlighted in bold.}
    \label{tab:results}
    \begin{tabular}{lccccc}
        \toprule
        Model & MRR & Precision & Recall & F1 Score & Accuracy \\ \midrule 
        \multicolumn{6}{c}{Competing models} \\ \midrule
        Ling-KNN & 0.33 & 0.248 & 0.247 & 0.248 & 0.989 \\
        Ling-SVM & 0.15 & 0.053 & 0.053 & 0.053 & 0.987 \\
        Ling-LR & 0.43 & 0.276 & 0.275 & 0.275 & 0.990 \\ \midrule
        \multicolumn{6}{c}{Proposed models} \\ \midrule
        Base model & 0.63 & 0.493 & 0.493 & 0.492 & 0.993 \\
        With anaphor text appended & 0.63 & 0.506 & 0.503 & 0.505 & 0.993 \\
        With candidate mask & 0.67 & 0.529 & 0.529 & 0.531 & 0.994 \\
        With agreement filtering & \textbf{0.81} & \textbf{0.706} & \textbf{0.706} & \textbf{0.709} & \textbf{0.996} \\
        \bottomrule
    \end{tabular}
\end{table*}

The results reported in Table \ref{tab:results} show that the sequence-to-sequence model we proposed for Arabic pronoun resolution, which does not require handcrafted linguistic features, outperforms the competing linguistics-based models in all evaluation metrics. The proposed base model achieved an MRR of 63\% and an accuracy of 99.3\%. It outperformed the best linguistic-based model by a margin of 20\% in MRR and 0.3\% in accuracy. However, one should keep in mind that since each example in the sequence is composed of one positive item (the antecedent) and many negatives (the remaining tokens), accuracy may not be adequate for assessing model performance due to the unbalanced data nature of the sequence. In terms of precision and F1 score, the proposed model outperforms the competing models by a margin of 22\%. These results demonstrate that our proposed base model can effectively earn representations without relying on handcrafted linguistic features, thus enhancing Arabic pronoun resolution performance compared to competing models.
The results also show that the changes we introduced in the base model variants further improve its performance. Whereas appending the anaphor text shows only a minor increase in performance, masking non-candidate tokens showed improvements of 3\% and 2\% in MRR and F1 scores, respectively. Filtering candidate antecedents based on agreement resulted in the best performance among all models, with a 14\% improvement in MRR and an 18\% improvement in the F1 score compared to the base model. 

\begin{figure}[!t]
	\centering
	\subfloat[Case of attached pronoun]{\fbox{\includegraphics[width=\columnwidth]{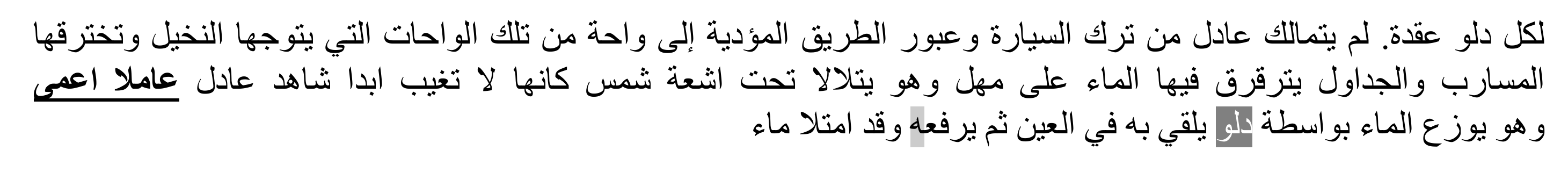}}%
		\label{fig:verbprocase}}
	
	\subfloat[Case of disagreement  between pronoun and the selected antecedent.]{\fbox{\includegraphics[width=\columnwidth]{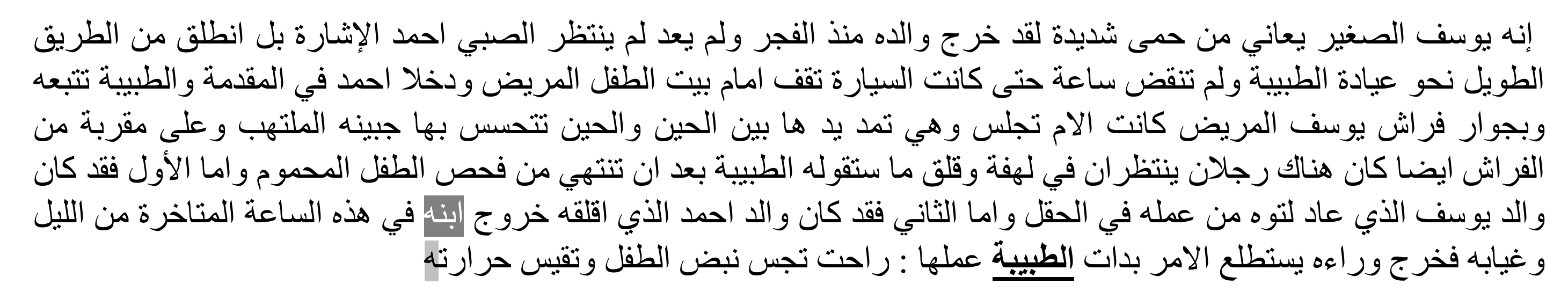}}%
		\label{fig:agreement}}
	
	\subfloat[Case of agreement  between pronoun and the selected antecedent.]{\fbox{\includegraphics[width=\columnwidth]{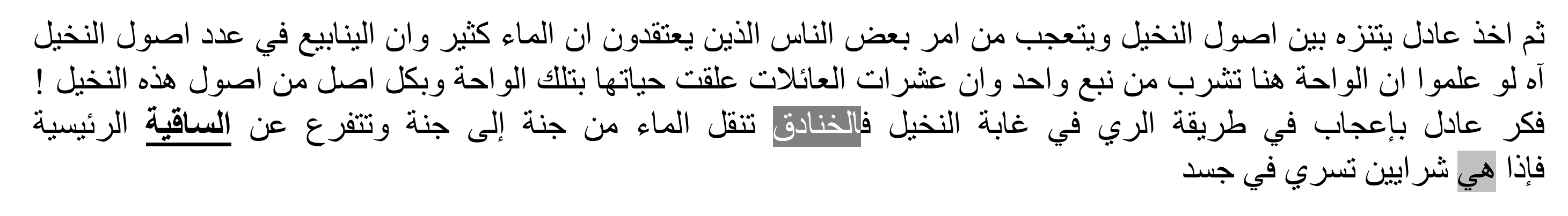}}%
		\label{fig:disagreement} }
	
	\subfloat[Case of a correct antecedent different from the annotated one.]{\fbox{\includegraphics[width=\columnwidth]{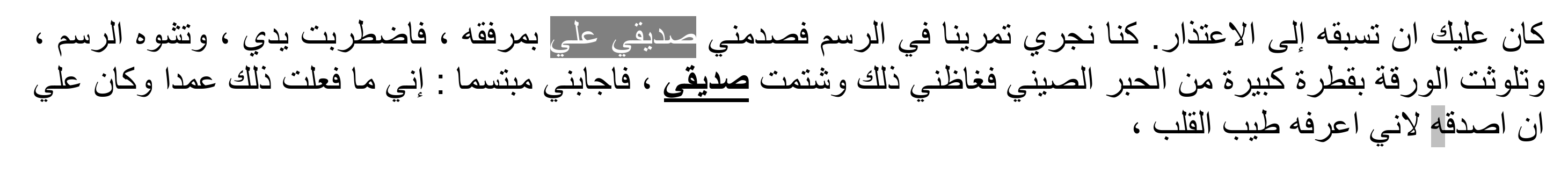}}%
		\label{fig:multimention} }
	
	\caption{Examples of false positive instances produced by the base model. The pronoun is highlighted in light gray, the correct antecedent is highlighted in dark gray, and the predicted antecedent is underlined.}
\end{figure}

To gain a deeper understanding of the proposed model, we randomly selected 100 instances of the 242 false positives produced by the base model and analyzed the model's selection behavior. One primary source of error we found is related to attached pronouns. In Arabic, an attached pronoun poses some ambiguity when attached to a verb. In this situation, the whole verb, including the pronoun, refers to the subject, while the pronoun part refers to the object. Consequently, the model may occasionally select the subject instead of the object, as illustrated in Figure \ref{fig:verbprocase}.

We also discovered that 15\% of selected instances resulted from typographical errors in the original text, mistakes in annotations that were not identified during the preprocessing stage, inaccurately selected antecedents, or the model selecting a correct mention that differed from the annotated one. As shown in Figure \ref{fig:multimention}, in some instances, the antecedent was mentioned in multiple ways, and the annotation identified the first mention, while the model selected the nearest one, which was correct but not the same as the annotation. This observation highlights the need for clear guidelines in the annotation process for Anaphora Resolution tasks, particularly for confusion cases involving attached pronouns. Moreover, it emphasizes the need to establish a consistent annotation method, such as selecting the first mention or the last, to facilitate evaluating and comparing anaphora resolution models.

\begin{figure}[!t]
	\centering
	\subfloat[Scores obtained by the base model.]{\fbox{\includegraphics[width=\columnwidth]{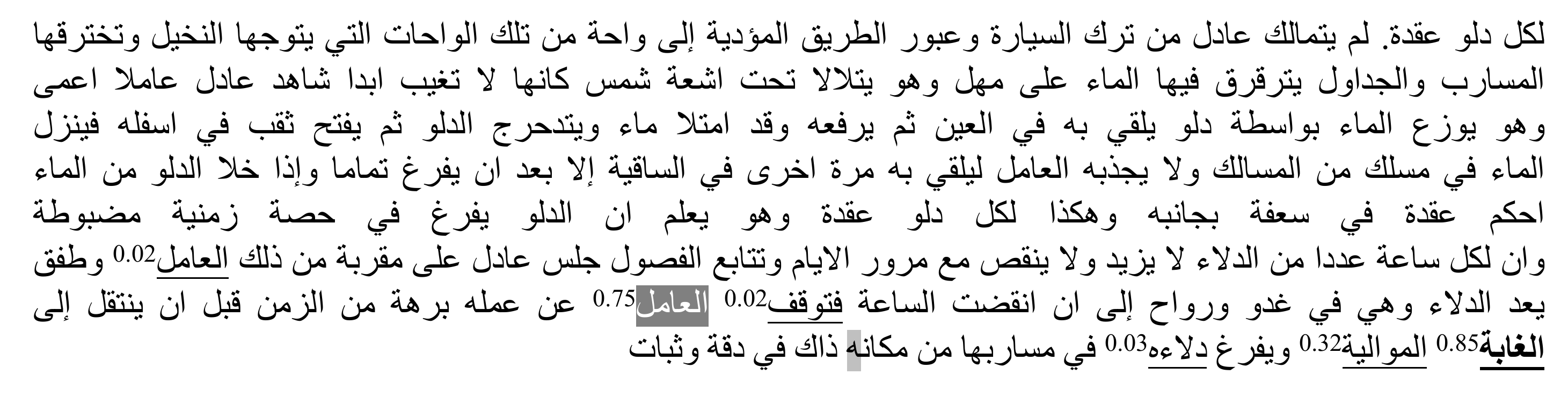}}
		\label{fig:basescore}}
	
	\subfloat[Scores after applying the candidate mask.]{\fbox{\includegraphics[width=\columnwidth]{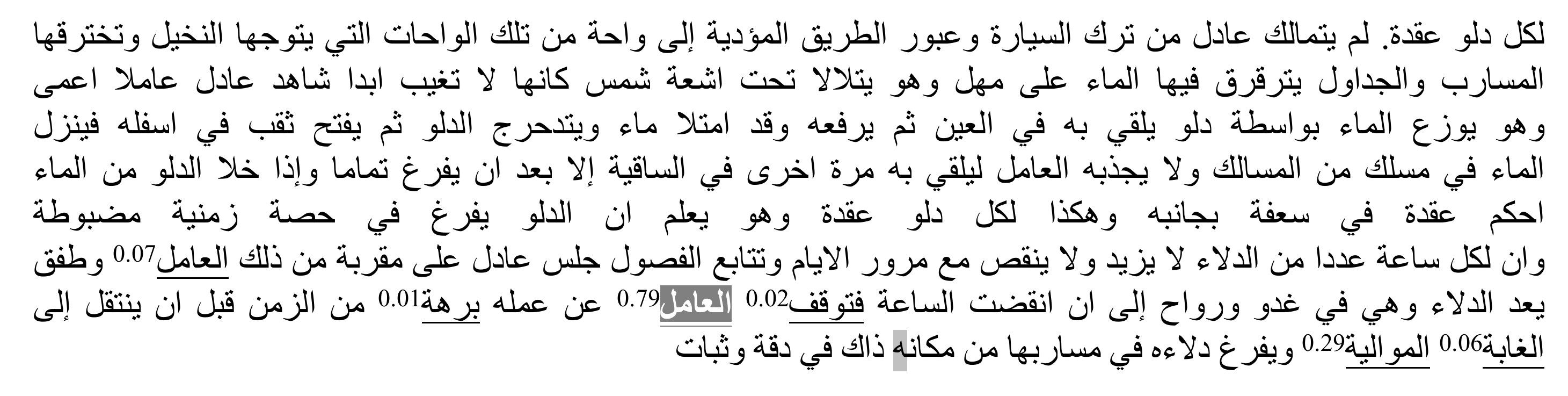}}
		\label{fig:maskscore}}
	\caption{The effect of candidate mask on the produced scores. We use light gray for the pronoun, dark gray for the correct antecedent, bold for the predicted antecedent, and underline for top scored candidates.}
	\label{fig:score}
\end{figure}

Using the candidate mask led to a 20\% improvement in correctly identifying the chosen antecedent in the selected instances, thereby reducing the number of false positives. It is worth noting, however, that the mask does not eliminate a false positive by zeroing the selected antecedent's score since the latter is already on the candidate list. Instead, it indirectly alters the scores assigned to the candidates, as depicted in Figure \ref{fig:score}. In the base model, the correct antecedent "\RL{العامل}" (the worker), was ranked second, whereas the word "\RL{الغابة}" (the forest)  is wrongly selected as the antecedent. In this case, the predicted antecedent "\RL{الغابة}" is actually related to the noun part of the (noun + attached pronoun) construction, specifically to "\RL{مكانـ}" (place). This ambiguity in the attached pronoun represents a challenge for the model's understanding. Nevertheless, using the candidate mask after the sigmoid layer resolves the ambiguity in this example, as shown in Figure \ref{fig:maskscore}, even though the candidate "\RL{الغابة}" is still assigned a nonzero score.

Many unresolved cases are caused by disagreements in number or gender and require certain reasoning to determine the correct antecedent between the actual value on annotation or the predicted one, as shown in Figure \ref{fig:disagreement}. For instance, the attached pronoun "his," being masculine singular, was predicted to reference the "Doctor" (a female doctor), resulting in a gender disagreement. In addition, Arabic has a specific type of anaphora references incompatible in gender or number as exemplified in Figure \ref{fig:agreement}. In this case, the model could not process the feminine singular pronoun used in a plural context, which resulted in selecting a candidate that agreed with the gender and number requirements. 

A minor number of gender-number agreement errors were resolved upon using the candidate mask, and the model was still unable to capture this type of information due to the limited data or the limitations of the Bi-LSTM architecture, or the pre-trained Arabert. Filtering the candidates to include only those that agree with the number or gender requirements and have at least two candidates significantly improved the model's performance. Among the selected examples, the resolved cases increased by 32\% compared to using the candidate mask only, reaching a total of 53\%. However, there are still mistakes remaining due to incorrect information about gender or number. For instance, the word "\RL{عينيه}" (his eyes)  is a dual form, but the analyzer incorrectly identifies it as singular, resulting in an incorrect result. Moreover, when the candidate is a phrase, the analyzer may not accurately identify the number or gender information of the phrase as a whole. Additionally, it is worth noting that annotation errors still exist, and mistakes in annotation and offset mapping continue to cause false positive examples.

With Arabic's complex morphological and syntactic structure, it has been observed that the proposed model, including AraBERT and Bi-LSTM can learn representations automatically, does not improve performance in certain linguistic situations. These situations include cases with ambiguous references to pronouns or when the agreement between different sentence parts is not straightforward. Despite these limitations, the proposed model has demonstrated its potential in resolving pronouns in unambiguous instances, achieving a success rate of around 76\% out of approximately 500 examples. It is worth noting that in cases of false positives, the model did not produce meaningless results but rather frequently made selections that closely approximated the correct antecedent or had some relation to the pronoun or its phrase.

\section{Conclusion}
This work investigates the effectiveness of deep learning-based NLP techniques, including Bi-LSTM, and pre-trained language models like BERT, for Arabic anaphora resolution. The proposed model outperformed the baselines linguistics KNN, SVM, and Logistic Regression models across all evaluation metrics, achieving a 43\% improvement in MRR and higher precision, F1 score, and accuracy.

We gave a detailed description of the model architecture and several modifications to improve its performance. 
The modification process followed a systematic and iterative approach, wherein each improvement was built upon the insights acquired from previous experiments. The initial model achieved a 63\% MRR and 49\% F1 score.
The first step to improve it was incorporating the anaphor text into the sequence, which resulted in minor improvements.
Drawing from the insights gained through this initial modification, we introduced a layer that masked candidates' scores. 
We then further refined the model by modifying the candidate list and introducing more stringent filters, including considerations for number and gender agreement. This iterative improvement process significantly boosted the model's performance, culminating in an 81\% MRR and 71\% F1 score.

This work contributes to growing the body of research on anaphora resolution in Arabic, highlighting the potential of deep learning techniques to enhance the state-of-the-art in this field. Our work emphasizes the importance of investigating the transferability and generalizability of NLP techniques for Arabic. 
There is a clear need to create comprehensive annotated datasets considering the diversity of Arabic pronouns. More efforts are needed to devise clear annotation guidelines, especially given the complex morphological features of Arabic, such as attached pronouns. Such robust guidelines would not only enhance our current models but also pave the way for future research in Arabic anaphora resolution.

Whereas this paper focuses primarily on Arabic pronoun resolution, the insights and methodology presented can be applied to other references within the text, such as cataphora, zero-anaphora, and bridging anaphora. We believe that this research provides a solid foundation for future studies in this area, and we hope that our findings encourage researchers to explore the potential of deep learning techniques in addressing a wide range of Arabic linguistic tasks.

\section*{Acknowledgment}
This research project was supported by a grant from the “Research Center of the Female Scientific and Medical Colleges”, Deanship of Scientific Research, King Saud University.

\IEEEtriggeratref{32}
\bibliographystyle{IEEEtran}
%\bibliography{bibliography}
% Generated by IEEEtran.bst, version: 1.14 (2015/08/26)

\end{document}